\documentclass[letterpaper, 10 pt, conference]{ieeeconf}  

\IEEEoverridecommandlockouts                              

\usepackage{censor}
\usepackage{graphicx}
\usepackage{amsmath}
\usepackage{amssymb}
\usepackage{xcolor}
\usepackage{booktabs}
\usepackage{subcaption}
\usepackage{url,cite} 
\usepackage{siunitx}
\usepackage{multirow}
\usepackage{tikz}
\usetikzlibrary{arrows.meta,positioning,calc,decorations.pathreplacing}
\usepackage{pgfplots}
\pgfplotsset{compat=1.18}

\title{\LARGE \bf
Mind the Shape Gap: A Benchmark and Baseline for Deformation-Aware 6D Pose Estimation of Agricultural Produce
}

\author{N. Chatzis$^{*,1,2,3}$, A. Tsinouka$^{*,1,2}$, K. Papadimitriou$^{1,2,5}$, N. Efthymiou$^{1,2}$, M. Glytsos$^{4}$, \\ G. Retsinas$^{*,1}$, P. Oikonomou$^{1,2,3}$,  G. Potamianos$^{1,2,5}$, P. Maragos$^{1,2}$, and P. P. Filntisis$^{1,2}$%
\thanks{$^{*}$Equal Contribution.}
\thanks{$^{1}$Robotics Institute, Athena Research Center, Marousi, Greece}%
\thanks{$^{2}$HERON - Hellenic Robotics Center of Excellence, Athens, Greece}%
\thanks{$^{3}$School of Electrical \& Computer Engineering, NTUA, Greece}%
\thanks{$^{4}$Music Technology, New York University, USA}
\thanks{$^{5}$Department of 
Electrical \& Computer Engineering, UTH, Greece}
}

\begin{document}

\maketitle
\thispagestyle{empty}
\pagestyle{empty}

\begin{abstract}
Accurate 6D pose estimation for robotic harvesting is fundamentally 
hindered by the biological deformability and high intra-class shape 
variability of agricultural produce. Instance-level methods fail in 
this setting, as obtaining exact 3D models for every unique piece of 
produce is practically infeasible, while category-level approaches 
that rely on a fixed template suffer significant accuracy degradation when 
the prior deviates from the true instance geometry. To bridge such lack of robustness to deformation, we introduce PEAR (Pose and 
dEformation of Agricultural pRoduce), the 
first benchmark providing joint 6D pose and per-instance 3D 
deformation ground truth across 8 produce categories, acquired via a 
robotic manipulator for high annotation accuracy. Using 
PEAR, we show that state-of-the-art methods suffer up to $6\times$ 
performance degradation when faced with the inherent geometric deviations of real-world produce. 
Motivated by this finding, we propose SEED (Simultaneous Estimation 
of posE and Deformation), a unified RGB-only framework that jointly 
predicts 6D pose and explicit lattice deformations from a single 
image across multiple produce categories. Trained entirely on 
synthetic data with generative texture augmentation applied at the UV 
level, SEED outperforms MegaPose on 6 out of 8 categories under 
identical RGB-only conditions, demonstrating that explicit shape 
modeling is a critical step toward reliable pose estimation in 
agricultural robotics.



\end{abstract}

\section{INTRODUCTION}

Robotic systems offer a scalable route to automate harvesting and field operations, easing labor constraints, improving throughput, and helping agriculture meet rising global food demand~\cite{kootstra2021selective,zhang2022algorithm}. 
However, large-scale deployment in the wild requires robustness across a wide spectrum of perception and manipulation challenges. Vision algorithms must remain robust to dynamic outdoor illumination and unpredictable weather, while agricultural scenes introduce severe occlusions from dense foliage and unstructured clutter, heavily restricting available visual information.
Beyond these scene-wide difficulties, the agricultural produce itself exhibits substantial intra-class diversity in shape, size, and surface texture, all of which evolve throughout the crop lifecycle, making accurate 3D modeling for reliable robotic grasping exceptionally difficult.

\begin{figure}[!htbp]
    \centering
    \includegraphics[width=1.0\linewidth]{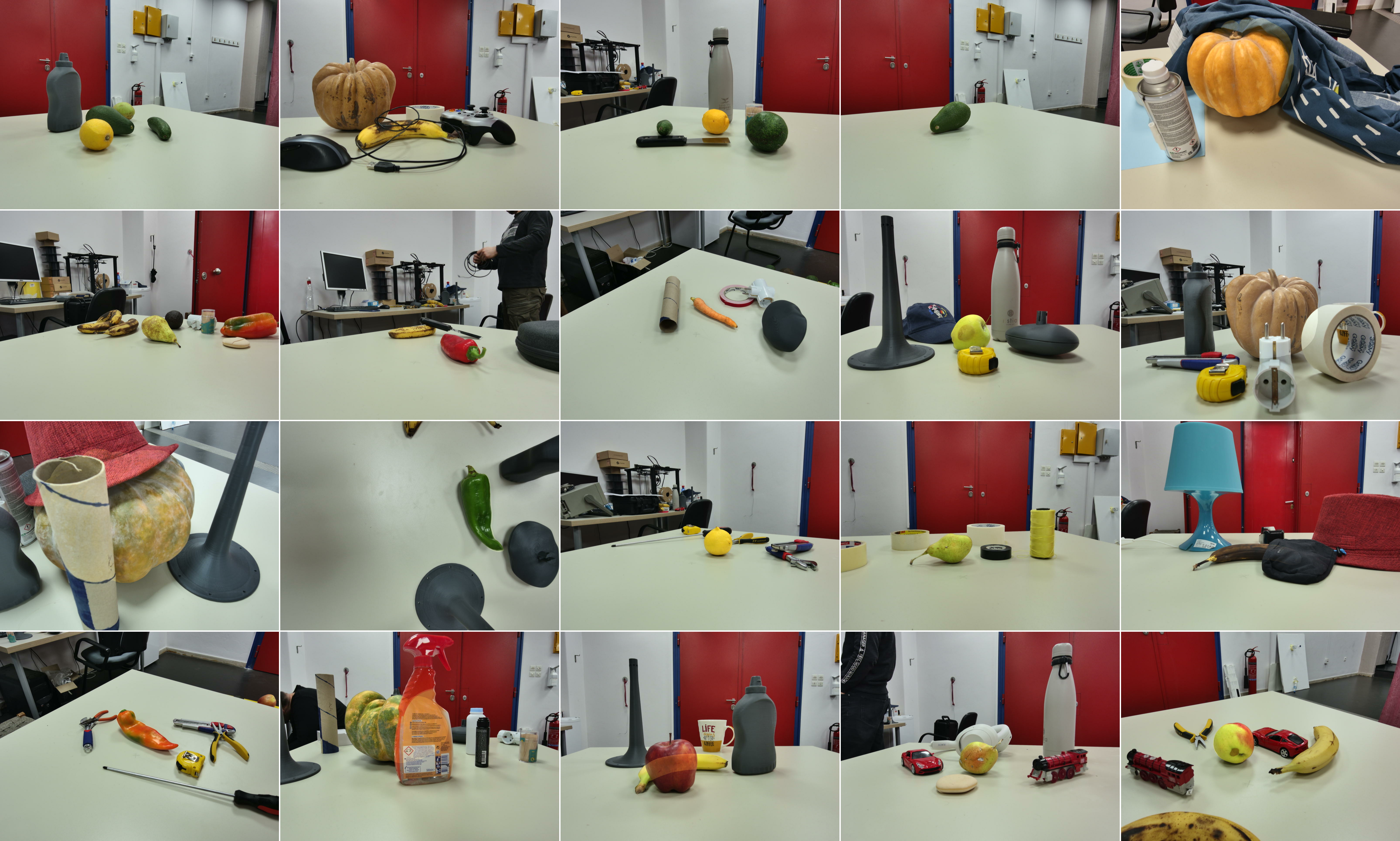}
    \caption{Sample annotated frames from the PEAR dataset.}
    \label{fig:dataset}
    \vspace{-0.5cm}
\end{figure}

Currently, existing 6D pose estimation methods typically rely on knowing the specific object instance beforehand. Whether utilizing purely RGB inputs, such as CozyPose~\cite{cosypose}, MegaPose~\cite{labbe2022megapose}, and RGBTrack~\cite{RGBTrack}, or leveraging depth cues~\cite{foundation_pose, any6d}, these instance-level approaches require exact 3D CAD models, or multiple reference images to build a representation of the target's 3D shape at inference time. While highly effective in controlled laboratory setups, these instance-specific requirements fail in challenging agricultural environments, where obtaining exact models for every unique piece of produce or capturing multiple clear views on-the-fly is practically infeasible.

An alternative is category-level pose estimation, which bypasses instance-specific requirements by generalizing to unseen objects within a known class~\cite{omnionocs,eif6d,catdeform,yopo}. While recent methods efficiently model complex intra-class shape variations, the most robust among them heavily depend on depth sensors (RGB-D)~\cite{socs,eif6d,catdeform}, making them highly sensitive in outdoor settings. Conversely, RGB-only approaches struggle to efficiently capture the geometric structure of specific produce, such as the inherent stretched curvature of a banana or the approximate symmetry of a tomato. A critical open question is therefore: \emph{how severely does the shape mismatch between a category-level template and the actual instance geometry affect pose estimation in practice?}

\begin{figure}[!htbp]
    \centering
    \includegraphics[width=1.0\linewidth]{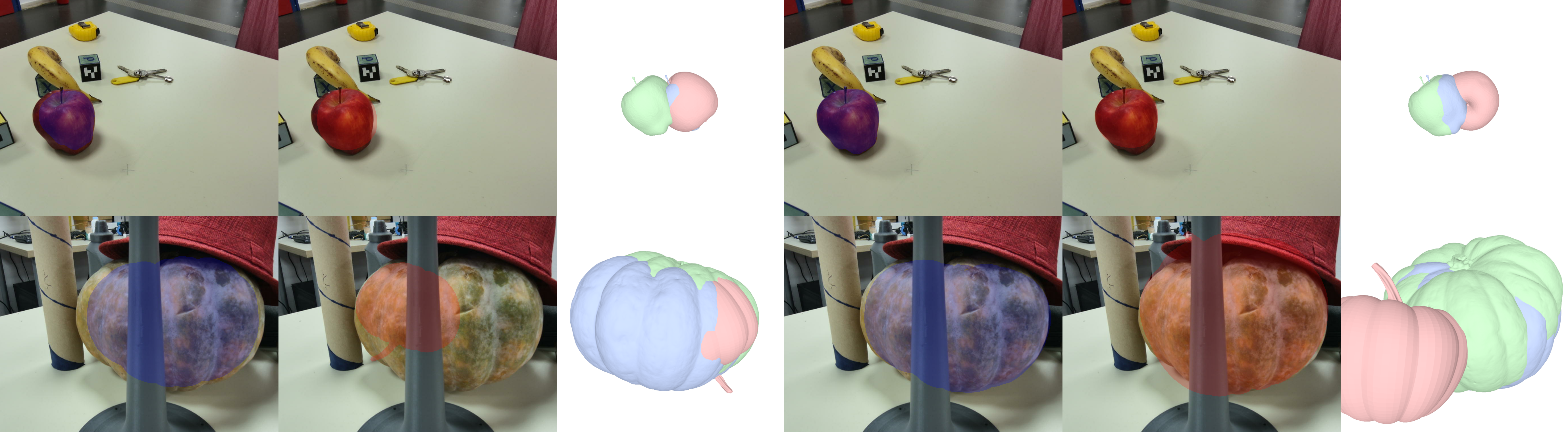}
    \caption{Effect of template shape mismatch on pose estimation. Each row shows the results of foundation models for a selected frame of PEAR dataset. From left to right: FoundationPose predictions using the ground-truth mesh, predictions using a category-level base mesh, and a reference view for comparison; the same three visualizations are shown for MegaPose in the last three columns. The results highlight the sensitivity of pose estimators to geometric discrepancies between the template mesh and the true object shape.}
    \label{fig:theproblemoffoundationmodels}
    \vspace{-0.3cm}
\end{figure}

However, answering this question has been fundamentally hindered by a severe lack of suitable benchmarks. Existing agricultural datasets~\cite{deepfruit,Fuji-SfM} focus primarily on 2D tasks or assume rigid templates, and completely lack joint instance-specific 6D pose and explicit 3D deformation annotations. To bridge this gap, we introduce the PEAR benchmark, a unified evaluation framework designed specifically for joint 6D pose and 3D shape estimation across 8 produce categories. Using PEAR, we provide the first controlled study of how per-instance geometric deformation impacts state-of-the-art pose estimators (Fig. ~\ref{fig:theproblemoffoundationmodels}), revealing performance degradations of up to $6\times$ when methods lack access to the true instance geometry.

Motivated by this finding, we further ask whether explicit deformation modeling can recover the lost accuracy. By leveraging class-specific geometric priors, recent work like PLANTPose~\cite{glytsos2025category} demonstrated that extreme intra-class variation can be effectively modeled from RGB inputs through lattice-based deformations of a base mesh. However, this architecture is inherently restricted to a single produce category, limiting its practical scalability. Building upon the conceptual foundation of PLANTPose, we introduce SEED, which extends deformation modeling to multiple agrifood categories and demonstrates that explicit shape prediction can narrow the deformation gap in a purely RGB setting.

Our core contributions are twofold:
\begin{itemize}
    \item We introduce PEAR, a novel benchmark providing precise 6D pose and 3D deformation annotations across 8 produce categories, featuring 5 shape and texture variations per class, acquired via a robotic manipulator to ensure high annotation accuracy. PEAR includes real evaluation sequences and a large-scale synthetic 
training set, providing the first controlled evidence that per-instance 
shape deformation is a critical bottleneck--with state-of-the-art 
methods, as demonstrated by their up to $6\times$ degradation, when lacking true instance geometry.
    \item We propose SEED, a unified architecture for simultaneous multi-category 6D pose and explicit deformation estimation from single RGB images. Trained entirely on the PEAR synthetic data with generative AI texture augmentation applied at the UV level, SEED outperforms MegaPose on 6 out of 8 categories under identical RGB-only conditions, demonstrating that explicit deformation modeling is a promising direction for closing the gap toward instance-level accuracy.
\end{itemize}



\section{Related Work}
\subsection{Instance-level 6D Pose Estimation}
Instance-level methods require prior knowledge of an object's exact 3D geometry, and fall into two categories: RGB-only and RGB-D approaches. RGB-only instance-level models have evolved from classical regression and correspondence pipelines, such as CosyPose~\cite{cosypose} and GPose~\cite{GPose}, to more flexible zero-shot frameworks. Recent advancements like MegaPose~\cite{labbe2022megapose}, GigaPose~\cite{nguyen2024gigapose}, and Co-op~\cite{coop} enable the handling of novel items without retraining by matching query images to pre-rendered templates or foundation model features. Furthermore, recent frameworks like RGBTrack~\cite{RGBTrack} rely on true-scale 3D models. On the other hand, RGB-D approaches leverage depth cues to achieve higher precision and robust metric-scale estimation. Methods like FoundationPose~\cite{foundation_pose} bridge the gap between model-based and model-free setups by learning neural implicit representations from either CAD models or a few reference views. Similarly, Any6D~\cite{any6d} simplifies the process by requiring only a single RGB-D anchor image as a reference for pose inference. PartPose~\cite{partpose} addresses multi-part objects by modeling them as kinematic chains of rigid links, utilizing Bayesian optimization to refine the pose of articulated components.

Unfortunately, such instance-specific requirements, ranging from exact CAD models to rigid-part assumptions, are highly restrictive for agricultural applications. The inherent organic deformability and geometric variability of crops render rigid-template-based pose estimation practically infeasible for unstructured field environments.

\subsection{Category-Level Pose Estimation}
To bypass the need for exact instance models, category-level methods predict the pose of unseen instances within a known class. In the RGB-only domain, DISP6D~\cite{disp6d} disentangles shape and pose latent spaces, while OmniNOCS~\cite{omnionocs} scales this paradigm to ``in-the-wild'' scenarios across numerous categories using deformable canonical models. More recently, YOPO~\cite{yopo} utilizes a Transformer architecture to directly predict poses and scales. In the context of agricultural robotics, PLANTPose~\cite{glytsos2025category} has been developed to handle the specific geometry of bananas, leveraging category-level priors for 6D pose estimation.

Incorporating depth at inference allows for more accurate modeling of non-rigid shape deformations from categorical priors. SOCS~\cite{socs} utilizes 3D thin-plate spline warping for smooth non-rigid deformation, while ShAPO~\cite{shapo} leverages implicit fields for joint shape and pose prediction. Recent state-of-the-art methods like CatDeform~\cite{catdeform} and EIF-6D~\cite{eif6d} explicitly warp prior point clouds using Transformer attention modules and dual-stream networks, respectively, to handle severe non-rigid deformations and partial occlusions.

\subsection{Synthetic Data for 6D Pose Estimation}
Many of the aforementioned works heavily rely on synthetic data, due to the difficulty of annotating real images. Most methods leverage mesh models from large-scale 3D databases, such as ShapeNet~\cite{shapenet}, Google Scanned Objects~\cite{google_scanned_objects}, and OmniObject3D~\cite{omnionocs}, rendering them in pipelines like BlenderProc~\cite{blenderproc}. To bridge the sim-to-real gap, diffusion-based approaches have shown great promise. For instance, PLANTPose~\cite{glytsos2025category} introduces a generation pipeline that combines physics-based simulation with image enhancement via image-to-image Stable Diffusion inpainting, creating highly realistic synthetic training data. However, while synthetic data are crucial for training, validating these models requires high-fidelity real-world benchmarks that capture both the geometric complexity and the organic deformation of agricultural produce.

\subsection{6D Pose Estimation in Agriculture}
The complexities of robotic harvesting have driven the development of domain-specific pose estimation pipelines. The work in~\cite{retsinas2023mushroom} uses implicit pose encodings for mushroom harvesting, while Deep-ToMaToS~\cite{KIM2022107300} simultaneously predicts tomato ripeness and 6D pose. More recently, TomatoPoseNet~\cite{tomatoposenet} has introduced a keypoint-based model, utilizing multiscale feature fusion to estimate the 6D pose of cutting points on fruits. Similarly, Li et al.~\cite{10802107} proposed a method for strawberry 6D pose and 3D scale estimation, and Costanzo et al.~\cite{costanzo2023enhanced} presented an enhanced RGB-D system for apple grasping. PLANTPose~\cite{glytsos2025category} recently demonstrated the viability of lattice-based deformations for a single produce category from purely RGB images. However, its architecture is limited to a single class, and its reliance on image-level generative inpainting inadvertently alters the underlying surface geometry, corrupting the synthetic ground-truth. These limitations highlight a clear gap in the literature, which SEED addresses via a category-level RGB-only pipeline, realistic augmentation, and a purpose-built agricultural benchmark.

\section{The PEAR Benchmark}
In this section we describe the creation of the PEAR (\textbf{P}ose and d\textbf{E}formation of \textbf{A}gricultural p\textbf{R}oduce) benchmark (Fig.~\ref{fig:dataset}), which includes 3D scans of multiple produce instances across 8 agrifood categories, paired with multi-view RGB-D sequences, captured via a robotic manipulator to provide accurate ground-truth annotations for 6D pose under varying levels of occlusion.

Annotating 6D object poses in real data is inherently difficult: limitations in image 
resolution, sensor noise, and viewpoint ambiguity can significantly degrade single-frame pose estimates, making it difficult to obtain trustworthy annotations. This problem becomes substantially more severe under occlusion, where large portions of the object may be invisible to the camera. To address this, our acquisition and annotation pipeline 
consists of three stages: \textit{(i)} agrifood collection and 3D scanning; \textit{(ii)} multi-view scene capture using a robotic manipulator; and \textit{(iii)} automatic pose 
annotation via multi-view optimization with robust inlier selection. We describe each stage in detail below.

\subsection{Object Collection and Scanning}
\label{subsec:scanning}

We collect 5 instances across 8 produce categories: 
banana, pumpkin, pear, apple, lemon, long pepper, carrot, and avocado, selecting diverse instances that cover a representative range of shape and 
texture variations within each category.
We deliberately choose categories whose shape variations are smooth and global — a banana that curves more or less, a pepper that stretches or tapers — rather than structurally disjoint ones like grape clusters, whose branching topology cannot be meaningfully captured by a simple lattice-based warp.

Each instance is 3D scanned using a Creality Raptor Pro scanner\cite{creality_raptor_pro} ($\approx \qty{0.02}{\milli\meter}$ accuracy), yielding 
high-fidelity meshes that serve as the ground-truth 
geometry.


\subsection{Multi-View Capture with Franka Panda}
\label{subsec:capture}

To obtain precise and repeatable camera extrinsics, we 
rigidly mount an Orbbec Femto Bolt RGB-D camera\cite{orbbec_femto_bolt} onto a 
Franka Emika Panda robot\cite{franka_panda}. The robot's kinematic model 
provides camera poses with sub-millimeter translational 
and sub-degree rotational accuracy, allowing us to treat 
the extrinsics as ground truth. The robot is commanded 
to stop fully at each target pose before image acquisition 
to eliminate motion blur and ensure measurement stability. Fig.~\ref{fig:protocol_setup} shows our setup.

\subsubsection{Camera Placement Strategy}
\begin{figure}[t]
\centering

\begin{subfigure}{0.30\textwidth}
\centering
\includegraphics[width=1\textwidth]{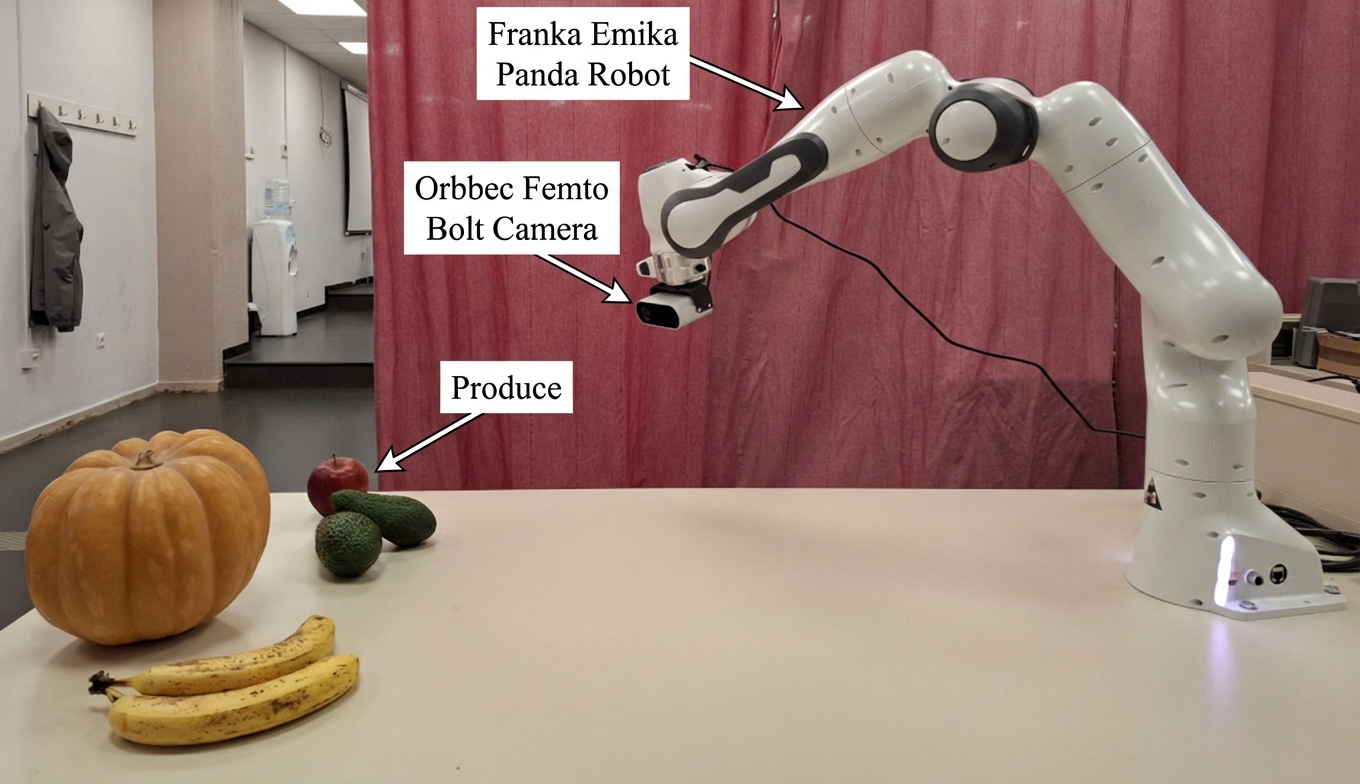}
\caption{Robotic acquisition setup.}
\label{fig:protocol_setup}
\end{subfigure}
\hfill
\begin{subfigure}{0.30\textwidth}
\centering
\begin{tikzpicture}[
  x=0.95cm,y=0.95cm, line cap=round, line join=round, >=Latex,
  every node/.style={font=\tiny},
  ring/.style={thick, draw=blue!40},
  ray/.style={-Latex, thin, gray!80},
  dashedthin/.style={densely dashed, thin, gray},
  box/.style={draw=gray!50, rounded corners, fill=gray!5, inner sep=3pt, align=left, font=\tiny}
]
\coordinate (C) at (0,0);
\fill[red] (C) circle (1.5pt);
\node[right, font=\tiny] at (C) {Produce};

\draw[ring] (C) circle (0.9);
\draw[ring] (C) circle (1.3);
\draw[ring] (C) circle (1.7);

\node[font=\tiny, text=blue!80!black] (L50) at (0.2, 1.8) {$50$cm};
\node[font=\tiny, text=blue!80!black] (L60) at (0.9, 1.8) {$60$cm};
\node[font=\tiny, text=blue!80!black] (L70) at (1.6, 1.8) {$70$cm};

\draw[->, thin, blue!80!black] (L50) -- ($(C)+(90:0.9)$);
\draw[->, thin, blue!80!black] (L60) -- ($(C)+(70:1.3)$);
\draw[->, thin, blue!80!black] (L70) -- ($(C)+(50:1.7)$);

\draw[dashedthin] (-2.2, 0) -- (C);
\fill[gray!40] (-2.2,-0.12) rectangle (-2.0, 0.12);
\node[left, font=\tiny] at (-2.2, 0) {Base};

\def\r{1.7}
\draw[line width=1.2pt, blue!80!black] ($(C)+(140:\r)$) arc[start angle=140,end angle=220,radius=\r];

\foreach \a in {140, 148, 156, 164, 172, 180, 188, 196, 204, 212, 220}{
  \fill[blue!80!black] ($(C)+(\a:0.9)$) circle (0.9pt);
  \fill[blue!80!black] ($(C)+(\a:1.3)$) circle (0.9pt);
  \fill[blue!80!black] ($(C)+(\a:1.7)$) circle (0.9pt);
}

\draw[ray] ($(C)+(140:1.6)$) -- ($(C)+(140:0.3)$);
\draw[ray] ($(C)+(220:1.6)$) -- ($(C)+(220:0.3)$);

\node[font=\tiny, text=blue!80!black] at (-1.4, 1.2) {$80^\circ$ arc};
\draw[->, thin, blue!80!black] (-1.4, 1.05) .. controls (-1.1, 0.85) .. (-1.05, 0.8);

\node[box, anchor=west] at (0.0, -0.9) {
  \textbf{Camera Viewpoints}\\
  $\bullet$ 3 Radii ($50, 60, 70$ cm)\\
  $\bullet$ 3 Elevations ($15^\circ, 30^\circ, 45^\circ$)\\
  $\bullet$ 11 Azimuths per arc\\
  $\bullet$ 3 Top-down ($80^\circ$ elev.)\\
  \textbf{Total:} 90 imgs/sequence
};
\end{tikzpicture}
\caption{Spherical camera viewpoint sampling.}
\label{fig:protocol_sampling}
\end{subfigure}
\caption{PEAR benchmark acquisition framework. (a) An Orbbec Femto Bolt RGB-D camera is rigidly mounted to a Franka Panda manipulator, observing the produce from a fixed distance. (b) To ensure comprehensive coverage, viewpoints are uniformly sampled across three concentric spheres along an $80^\circ$ azimuth arc facing the robot, always oriented directly toward the scene center.}
\label{fig:ripe_protocol}
\vspace{-0.6cm}
\end{figure}
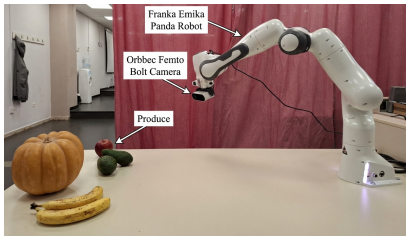
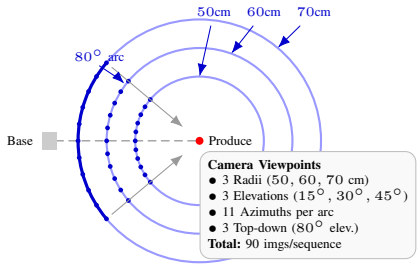

The scene center is fixed at \SI{80}{\centi\meter} in 
front of the robot base, where the object of interest 
is placed for all recordings. Camera viewpoints are 
sampled on three concentric spheres of radii 50, 60, 
and \SI{70}{\centi\meter}. On each sphere, lateral 
viewpoints are captured at three elevation angles 
($15^\circ$, $30^\circ$, $45^\circ$) by uniformly 
discretizing an $80^\circ$ polar arc into 11 
positions, centered on the robot-to-scene axis to 
avoid joint singularities. Three additional viewpoints 
are captured at $80^\circ$ elevation across a 
$20^\circ$ arc for each radius to provide top-down 
coverage. This yields a total of 90 camera placements 
per scene 
with the camera always oriented toward the 
scene center (see Fig.~\ref{fig:protocol_sampling}).

\subsubsection{Occlusion Protocol}
Scenes are recorded in paired sequences. First, the 
object is recorded across all 90 viewpoints in a clean, 
unoccluded configuration. Occluders are then introduced 
without disturbing the target object, and the robot 
revisits the exact same 90 poses. Since the object pose 
is established from the clean sequence, ground-truth 
annotations transfer directly to the occluded frames 
via the known extrinsics, enabling precise annotation 
even under severe visibility loss.

\subsection{Pose Annotation via Multi-View Optimization}
\label{subsec:annotation}

We next describe our optimization method. The key idea is to exploit the camera extrinsics to express all per-frame predictions in a common world coordinate frame. This unified representation enables global consistency enforcement during optimization and reliable inlier filtering afterwards.

\subsubsection{Camera Notation}
Let $W$ denote the world coordinate frame. For a 
recorded sequence of $N$ frames, let $C_i$ denote 
the camera frame at time $i$ and $O$ the object frame. 
The robotic extrinsics provide known rigid transformations:
\begin{equation}
\mathbf{T}^{W}_{C_i} \in SE(3), \quad i=1,\dots,N.
\end{equation}
For each frame, we estimate the object pose in camera 
coordinates $\hat{\mathbf{T}}^{C_i}_{O} \in SE(3)$, 
which induces a world-frame hypothesis:
\begin{equation}
\hat{\mathbf{T}}^{W}_{O,i} = \mathbf{T}^{W}_{C_i} \, 
\hat{\mathbf{T}}^{C_i}_{O},
\end{equation}
yielding the hypothesis set 
$\mathcal{D} = \{ \hat{\mathbf{T}}^{W}_{O,i} \}_{i=1}^N$.

\subsubsection{Joint Multi-View Optimization}

\begin{figure}[t]
    \centering
    \includegraphics[width=0.9\linewidth]{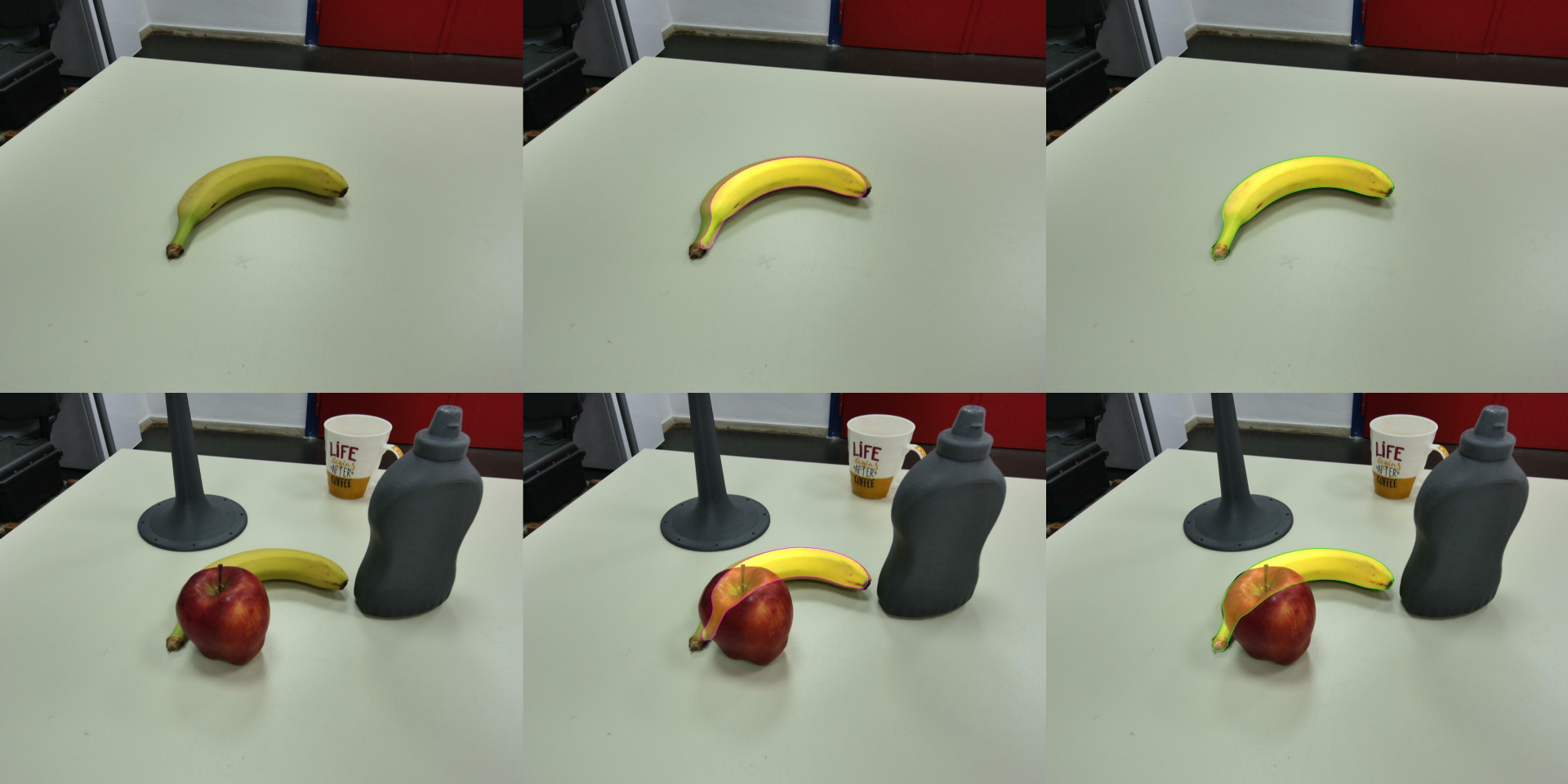}
    \caption{\textbf{Left}: input RGB images. \textbf{Middle}: initial pose predictions from FoundationPose given the depth image and object mask. \textbf{Right}: pose estimates after our joint multi-view optimization. The first row shows the object in an unoccluded setting, while the second row shows the same object placement under occlusion.}
    \label{fig:multiview_optimization}
    \vspace{-0.5cm}
\end{figure}

Our optimization scheme is designed to take initial pose estimates $\hat{\mathbf{T}}^{W}_{O,i}$ (denoted ${\mathbf{T}}_{i}$ for brevity), obtained independently per frame and refine them by improving per-frame alignment while enforcing agreement with a global multi-view consensus. In practice, poses are initialized using FoundationPose~\cite{foundation_pose} with object masks $\mathbf{M}_i$ from CNOS~\cite{nguyen2023cnos}, and then jointly refined by minimizing a total objective that combines a per-frame alignment loss and a global consistency term:
\begin{equation}
\mathcal{L}_{\text{total}} = \sum_{i=1}^{N} 
\mathcal{L}_{\text{align}}^{(i)} + \lambda \, 
\mathcal{L}_{\text{consist}},
\end{equation}
where $\lambda$ balances local image fidelity against 
global geometric consensus. A single-frame qualitative comparison is shown in Fig.~\ref{fig:multiview_optimization}, where our optimization improves the alignment between the rendered mesh and the observed image, compared to the initial estimation.

The per-frame alignment loss combines a perceptual 
term, a region overlap term, and a boundary alignment 
term:
\begin{equation}
\mathcal{L}_{\text{align}}^{(i)} =
w_{\text{VGG}}\,\mathcal{L}_{\text{VGG}}^{(i)}
+ w_{\text{dice}}\,\mathcal{L}_{\text{dice}}^{(i)}
+ w_{\text{dt}}\,\mathcal{L}_{\text{dt}}^{(i)},
\end{equation}
where $\mathcal{L}_{\text{VGG}}$ measures perceptual 
similarity between rendered and observed RGB within 
the object mask, $\mathcal{L}_{\text{dice}}$ measures 
region overlap between the rendered and observed 
silhouettes, and $\mathcal{L}_{\text{dt}}$ penalizes 
misalignment of rendered silhouette edges with respect 
to the ground-truth mask boundary via its distance 
transform $\mathbf{D}_i$ of the mask boundary. Formally, for an observed RGB image $\mathbf{I}_i$ and the rendered RGB $\hat{\mathbf{I}}_i$ coming from the mesh placed with the current pose:
\begin{align}
\mathcal{L}_{\text{vgg}}^{(i)} &= \sum_{\ell\in\mathcal{B}}
\frac{1}{|\Omega_\ell|}\left\|\phi_\ell(\hat{\mathbf{I}}_i)
-\phi_\ell(\mathbf{M}_i\odot\mathbf{I}_i)\right\|_1,\\
\mathcal{L}_{\text{dice}}^{(i)} &= 1-\frac{2\sum_{u}
\hat{\alpha}_i(u)\mathbf{M}_i(u)}
{\sum_{u}\hat{\alpha}_i(u)+\sum_{u}\mathbf{M}_i(u)
+\varepsilon},\\
\mathcal{L}_{\text{dt}}^{(i)} &= \frac{\sum_{u}
\|\nabla \hat{\alpha}_i(u)\|\,\mathbf{D}_i(u)}
{\sum_{u}\|\nabla \hat{\alpha}_i(u)\|+\varepsilon},
\end{align}
where $\phi_\ell$ denotes feature maps extracted from layer $\ell$ of a pretrained VGG network, $\Omega_\ell$ is the spatial domain of the corresponding feature map, $u$ indexes image pixels, $\hat{\alpha}_i \in [0,1]$ is the rendered soft object mask, and $\varepsilon$ is a small constant added for numerical stability.

The global consistency loss penalizes pairwise 
dispersion among world-frame hypotheses:
\begin{equation}
\mathcal{L}_{\text{consist}} = \frac{1}{|\mathcal{P}|} 
\sum_{(i,j)\in\mathcal{P}} \left( w_{r} \, 
e_R(\mathbf{T}_i, \mathbf{T}_j)^2 + w_{t} \, 
e_t(\mathbf{T}_i, \mathbf{T}_j)^2 \right),
\end{equation}
where $e_t$ denotes the Euclidean distance between translations of transformations and $e_R$ denotes the geodesic distance between rotations, evaluated over all unique frame pairs $ (i,j) \in \mathcal{P}$.

\subsubsection{Robust Inlier Selection}
Upon convergence, we apply a robust selection stage 
to reject any remaining tracking failures. Each 
hypothesis $\mathbf{T}_j \in \mathcal{D}$ is scored 
by the cardinality of its inlier support set:
\begin{equation}
\mathcal{I}_j = \left\{ i \;\middle|\; 
e_t(\mathbf{T}_i, \mathbf{T}_j) \leq \tau_t 
\;\land\; e_R(\mathbf{T}_i, \mathbf{T}_j) \leq 
\tau_R \right\},
\end{equation}
with $\tau_t = \SI{5}{\milli\meter}$ and 
$\tau_R = \SI{1}{\degree}$. The consensus pose is 
taken from the hypothesis maximizing 
$|\mathcal{I}_j|$, and all frames outside the 
winning inlier set are discarded as unreliable.
After removing outliers, a total of 4,500 annotations remained.


\begin{figure*}[t!]
    \centering
    \includegraphics[width=.84\textwidth]{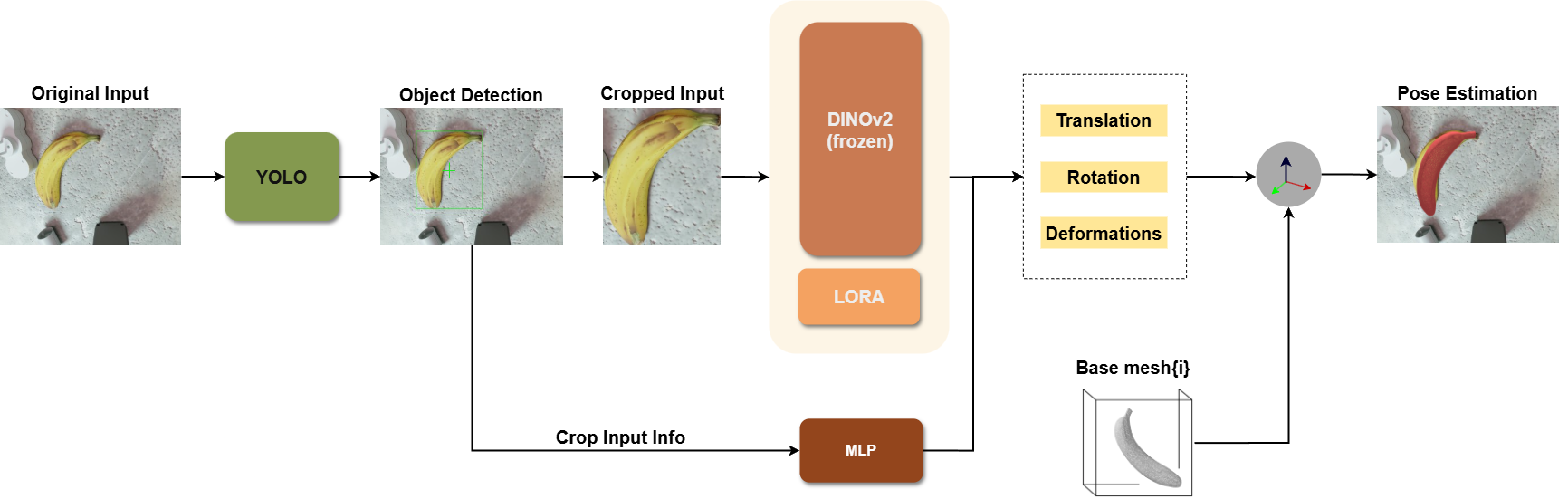}
    \caption{Overview of the SEED model. A cropped RGB detection is passed through a DINOv2 backbone with LoRA adaptation, producing features that feed three prediction heads for rotation, translation, and lattice deformation. The predicted deformation warps the category base mesh to match the observed instance geometry.}
\label{fig:model}
\vspace{-0.5cm}
\end{figure*}

\subsection{Synthetic Training Data Generation}
\label{subsec:synthetic_data}

Beyond real evaluation data, PEAR also provides a large-scale
synthetic training dataset with ground-truth 6D pose and per-instance
3D deformation labels, enabling researchers to train and benchmark
methods without requiring any real annotated images. The synthetic
generation pipeline comprises three stages: shape generation via
lattice deformations, realistic texture augmentation, and physically
plausible scene rendering.

\subsubsection{Shape Generation via Lattice Deformations}
We represent per-instance shape variation using the lattice
deformation scheme introduced in PLANTPose~\cite{glytsos2025category}.
For each of the 8 produce categories described in
Sec.~\ref{subsec:scanning}, a base mesh is collected from publicly
available 3D model repositories. A tight axis-aligned bounding box is
placed around each base mesh, and deformations are controlled by
displacing its 8 corners in 3D space, yielding a compact
24-dimensional deformation vector. Displacements are interpolated
across the mesh interior via cubic B-spline interpolation, which
guarantees $C^2$-continuity and produces smooth, realistic shape
variations. To disentangle deformation from pose, the Umeyama
algorithm~\cite{umeyama} is applied after each deformation to remove
any spurious rotation and translation introduced by the lattice
perturbation. For each category, random lattice deformations are
sampled within empirically determined bounds to generate a diverse set
of plausible shape instances. This 24-dimensional lattice
representation also serves as the canonical deformation
parameterization for any method trained or evaluated on PEAR.

\subsubsection{Texture Augmentation}
A key insight of our pipeline concerns the stage at which generative
augmentations are applied. In PLANTPose, Stable
Diffusion~\cite{rombach2022high} with
Control\-Net~\cite{zhang2023adding} conditioning was used to inpaint
the object directly on rendered images. While effective for bananas,
whose distinctive silhouette constrains the inpainting, we observed
that for other produce categories this image-level approach introduces
geometric artifacts: the inpainting can inadvertently alter the
apparent surface geometry, effectively corrupting the ground-truth
pose and deformation labels.

To avoid this issue, we instead apply generative augmentations at the
\emph{texture-map} level rather than the image level. Specifically, we
use Gemini 2.5 Flash Image~\cite{gemini_team_2025_gemini25} to edit the base UV texture
maps of each mesh, conditioned on prompts describing different
ripeness stages and visual appearances across the produce lifecycle.
This yields approximately 50 texture-map variants per category. Because
these edited textures are applied directly to the mesh UV layout
\emph{prior} to rendering, geometry annotations remain entirely
unaffected, producing significantly more consistent training data
across all categories.

\subsubsection{Scene Rendering}
Deformed and textured meshes are placed into synthetic scenes using
BlenderProc~\cite{blenderproc}, with procedural room geometry,
randomized lighting, and distractor objects to simulate the occlusion
and clutter typical of agricultural settings. Camera poses are sampled
around a point of interest, and images are rendered from multiple
viewpoints, yielding ground-truth annotations for 6D pose,
segmentation masks, and the 24-dimensional lattice deformation vector
for every object instance. 
Our final synthetic dataset comprises 1000 scenes for each instance rendered from 10 viewpoints each. Both the real evaluation and the synthetic dataset will be made publicly available to
facilitate future research in deformable object pose estimation for
agricultural robotics.

\begin{table*}[t]
\centering
\caption{Effect of shape deformation on pose estimation accuracy
(PEAR benchmark). FP = FoundationPose~\cite{foundation_pose},
MP = MegaPose~\cite{labbe2022megapose}. We report ADD, ADD-S, and
Chamfer distance (Ch.) in mm (lower is better). For each method we
compare performance using the ground-truth scanned mesh (GT) versus
the category-level base mesh (Base).}
\label{tab:benchmark_eval}
\resizebox{\textwidth}{!}{%
\setlength{\tabcolsep}{1.0pt}
\begin{tabular}{ll|ccc|ccc|ccc|ccc|ccc|ccc|ccc|ccc||ccc}
\toprule
& & \multicolumn{3}{c|}{\textbf{Banana}}
& \multicolumn{3}{c|}{\textbf{Pumpkin}}
& \multicolumn{3}{c|}{\textbf{Pear}}
& \multicolumn{3}{c|}{\textbf{Apple}}
& \multicolumn{3}{c|}{\textbf{Lemon}}
& \multicolumn{3}{c|}{\textbf{L.\ Pepper}}
& \multicolumn{3}{c|}{\textbf{Carrot}}
& \multicolumn{3}{c||}{\textbf{Avocado}}
& \multicolumn{3}{c}{\textbf{Average}} \\
\textbf{Method} & \textbf{Mesh}
& ADD & ADD-S & Ch.
& ADD & ADD-S & Ch.
& ADD & ADD-S & Ch.
& ADD & ADD-S & Ch.
& ADD & ADD-S & Ch.
& ADD & ADD-S & Ch.
& ADD & ADD-S & Ch.
& ADD & ADD-S & \multicolumn{1}{c||}{Ch.}
& ADD & ADD-S & Ch. \\
\midrule
\multirow{2}{*}{FP} & GT
& \textbf{36.1} & \textbf{17.6} & \textbf{17.6}
& \textbf{40.1} & \textbf{18.3} & \textbf{17.9}
& \textbf{33.5} & \textbf{17.7} & \textbf{17.6}
& \textbf{42.9} & \textbf{20.9} & \textbf{20.5}
& \textbf{43.3} & \textbf{22.5} & \textbf{22.2}
& \textbf{44.3} & \textbf{24.4} & \textbf{24.7}
& \textbf{35.5} & \textbf{22.6} & \textbf{22.9}
& \textbf{39.3} & \textbf{19.6} & \textbf{20.0}
& \textbf{39.4} & \textbf{20.4} & \textbf{20.4} \\
& Base
& 69.0 & 26.1 & 26.6
& 155.5 & 30.4 & 43.9
& 65.6 & 29.0 & 27.5
& 69.7 & 24.5 & 22.0
& 58.2 & 23.7 & 23.2
& 88.2 & 49.2 & 51.1
& 46.4 & 24.2 & 26.1
& 68.7 & 29.9 & 30.3
& 77.7 & 29.6 & 31.3 \\
\midrule
\multirow{2}{*}{MP} & GT
& \textbf{42.0} & \textbf{24.2} & \textbf{23.3}
& \textbf{100.0} & \textbf{11.7} & \textbf{11.3}
& \textbf{13.5} & \textbf{5.3} & \textbf{5.0}
& \textbf{20.6} & \textbf{3.6} & \textbf{3.4}
& \textbf{14.1} & \textbf{2.7} & \textbf{2.5}
& \textbf{14.8} & \textbf{7.8} & \textbf{7.7}
& \textbf{37.2} & \textbf{24.3} & \textbf{23.8}
& \textbf{26.6} & \textbf{8.2} & \textbf{7.9}
& \textbf{33.6} & \textbf{11.0} & \textbf{10.6} \\
& Base
& 97.3 & 62.4 & 61.3
& 223.0 & 93.9 & 104.3
& 110.8 & 72.9 & 72.7
& 72.6 & 29.4 & 28.3
& 92.1 & 59.0 & 57.9
& 109.7 & 74.3 & 77.3
& 107.1 & 85.3 & 85.7
& 91.7 & 55.6 & 53.3
& 113.0 & 66.6 & 67.6 \\
\bottomrule
\end{tabular}%
}
\vspace{-0.4cm}
\end{table*}

\section{Architecture for Joint 6D Pose \& Deformation Estimation}
\label{sec:method}

\subsection{Overview}
We now present SEED (Simultaneous Estimation of posE and Deformation),
a model for multi-category 6D pose and per-instance deformation
estimation from RGB images. SEED builds on the PLANTPose framework~\cite{glytsos2025category}, inheriting its lattice-based
deformation representation (Sec.~\ref{subsec:synthetic_data}), while
extending it with three key improvements: support for multiple object
categories, a stronger feature backbone, and a more robust training
procedure. An overview of the full pipeline is shown in
Fig.~\ref{fig:model}.

\subsection{Network Architecture}
\label{sec:network}

The input to SEED is a cropped RGB image of a single detected object
instance, obtained from a YOLOv11~\cite{yolo11_ultralytics} detector
trained on the PEAR synthetic data. We adopt a
DINOv2~\cite{oquab2024dinov2learningrobustvisual} Transformer backbone as our feature extractor.
The global image representation is formed by concatenating the CLS
token with the spatial average of all patch tokens. The backbone is
augmented with Low-Rank Adaptation
(LoRA)~\cite{hu2022lora} on all attention and MLP projections, which
efficiently fine-tunes the pretrained representations while keeping
the backbone weights frozen.

Since SEED operates on object crops, it lacks spatial context about absolute position and scale. We therefore encode a 4-dimensional crop placement vector $(c_x/W, c_y/H, w/W, h/H)$ through a small MLP and concatenate the result with the backbone features, where $c_x, c_y$ denote the crop center coordinates, $w, h$ the crop width and height, and $W,H$ the width and height of the original image. Three MLP heads (each 256 $\to$ 128 with batch normalization) are attached to the
fused representation:

\noindent\textbf{Rotation Head:} Predicts a 6D rotation
representation~\cite{zhou2019continuity}, orthonormalized into a valid
rotation matrix.

\noindent\textbf{Translation Head:} Outputs a 3D vector in
crop-relative coordinates (2D center offset and normalized depth),
converted to absolute camera-frame translation $(X, Y, Z)$ at
inference using the crop geometry and intrinsics.

\noindent\textbf{Deformation Head:} Predicts the 24-dimensional
lattice offset vector (see Sec.~\ref{subsec:synthetic_data}) that warps
the category base mesh to match the observed instance geometry.

\subsection{Training Losses}
\label{subsec:losses}

Let $\mathbf{R}$, $\mathbf{t}$, and $\boldsymbol{\delta}$ denote
the predicted rotation matrix, absolute translation, and lattice
offsets, respectively. We supervise training with six losses:

\noindent\textbf{Geodesic Rotation Loss.} We convert predicted and
ground-truth 6D representations to rotation matrices and penalize
their geodesic distance on $SO(3)$.

\noindent\textbf{Translation Loss.} MSE between predicted and
ground-truth translations, computed in meters:
$\mathcal{L}_{t} = \|\mathbf{t}^{pred} - \mathbf{t}^{gt}\|^2$.

\noindent\textbf{Deformation Loss.} MSE between predicted and
ground-truth lattice offsets:
$\mathcal{L}_{def} = \|\boldsymbol{\delta}^{pred} -
\boldsymbol{\delta}^{gt}\|^2$.

\noindent\textbf{Deformation Regularization.} An $L_2$ penalty on the
predicted offsets, biasing the network toward the undeformed base mesh.

\noindent\textbf{3D Vertex Loss.} The deformed mesh vertices are
transformed by the predicted and ground-truth poses and compared
directly in 3D camera coordinates. 


\noindent\textbf{2D Projection Loss.} The same 3D vertices are
projected onto the image plane using the camera intrinsics and
compared in normalized image coordinates.

The total training loss is:
\begin{multline}
    \mathcal{L} = \lambda_{rot}\mathcal{L}_{rot} +
    \lambda_{t}\mathcal{L}_{t} +
    \lambda_{def}\mathcal{L}_{def} \\
    + \lambda_{reg}\mathcal{L}_{reg} +
    \lambda_{3D}\mathcal{L}_{3D} +
    \lambda_{2D}\mathcal{L}_{2D}
\end{multline}

\noindent We train SEED end-to-end using AdamW with cosine annealing
on the PEAR synthetic training set (see Sec.~\ref{subsec:synthetic_data}) with standard image augmentations.

\begin{table*}[t]
\centering
\caption{SEED vs.\ MegaPose (Base mesh) on the PEAR benchmark —
both methods operate in a purely RGB setting with no per-instance
geometry. We report ADD, ADD-S, and Chamfer distance (Ch.) in mm
(lower is better). Best results in \textbf{bold}.}
\label{tab:seed_eval}
\resizebox{\textwidth}{!}{%
\setlength{\tabcolsep}{1.0pt}
\begin{tabular}{l|ccc|ccc|ccc|ccc|ccc|ccc|ccc|ccc|ccc}
\toprule
& \multicolumn{3}{c|}{\textbf{Banana}}
& \multicolumn{3}{c|}{\textbf{Pumpkin}}
& \multicolumn{3}{c|}{\textbf{Pear}}
& \multicolumn{3}{c|}{\textbf{Apple}}
& \multicolumn{3}{c|}{\textbf{Lemon}}
& \multicolumn{3}{c|}{\textbf{L.\ Pepper}}
& \multicolumn{3}{c|}{\textbf{Carrot}}
& \multicolumn{3}{c|}{\textbf{Avocado}}
& \multicolumn{3}{c}{\textbf{Average}} \\
\textbf{Method}
& ADD & ADD-S & Ch.
& ADD & ADD-S & Ch.
& ADD & ADD-S & Ch.
& ADD & ADD-S & Ch.
& ADD & ADD-S & Ch.
& ADD & ADD-S & Ch.
& ADD & ADD-S & Ch.
& ADD & ADD-S & Ch.
& ADD & ADD-S & Ch. \\
\midrule
MP-Base~\cite{labbe2022megapose}
& 97.3 & 62.4 & 61.3
& 223.0 & 93.9 & 104.3
& \textbf{110.8} & \textbf{72.9} & \textbf{72.7}
& 72.6 & 29.4 & 28.3
& \textbf{92.1} & \textbf{59.0} & \textbf{57.9}
& 109.7 & 74.3 & 77.3
& 107.1 & 85.3 & 85.7
& 91.7 & 55.6 & 53.3
& 113.0 & 66.6 & 67.6 \\
\textbf{SEED (Ours)}
& \textbf{67.4} & \textbf{41.6} & \textbf{42.4}
& \textbf{216.2} & \textbf{91.8} & \textbf{102.1}
& 124.5 & 89.7 & 92.2
& \textbf{62.0} & \textbf{23.6} & \textbf{22.9}
& 100.6 & 68.7 & 68.9
& \textbf{84.8} & \textbf{56.2} & \textbf{58.6}
& \textbf{78.3} & \textbf{60.1} & \textbf{60.6}
& \textbf{54.7} & \textbf{25.3} & \textbf{25.1}
& \textbf{98.6} & \textbf{57.1} & \textbf{59.1} \\
\bottomrule
\end{tabular}%
}
\vspace{-0.4cm}
\end{table*}

\section{Experimental Results}
\label{sec:experiments}

We structure our evaluation in two parts. First, we use the PEAR
benchmark to quantify the impact of geometric deformations on existing
state-of-the-art pose estimation methods, demonstrating that shape
variability is a critical and currently unaddressed bottleneck
(Sec.~\ref{subsec:benchmark_eval}). Second, we show how SEED can be
used to facilitate purely RGB-based pose and deformation estimation
across multiple produce categories
(Sec.~\ref{subsec:seed_eval}).

We report three metrics: ADD, the mean Euclidean distance between corresponding predicted and ground-truth mesh vertices; ADD-S, its symmetry-aware variant using closest-point matching; and Chamfer distance.

\subsection{Benchmarking the Effect of Shape Deformation}
\label{subsec:benchmark_eval}

To isolate the effect of per-instance shape variability on pose
estimation accuracy, we evaluate two leading methods on the PEAR
benchmark under two mesh conditions: using the \emph{ground-truth
scanned mesh} of each specific instance (GT), and using the
\emph{category-level base mesh} (Base) shared across all instances of
the same class. This controlled comparison directly reveals how much
performance degrades when the method lacks access to the true instance
geometry.

We evaluate MegaPose~\cite{labbe2022megapose} and
FoundationPose~\cite{foundation_pose} in the following
configurations:
\begin{itemize}
    \item \textbf{FP-GT}: FoundationPose with the ground-truth
    scanned mesh of each instance.
    \item \textbf{FP-Base}: FoundationPose with the category-level
    base mesh.
    \item \textbf{MP-GT}: MegaPose with the ground-truth scanned mesh.
    \item \textbf{MP-Base}: MegaPose with the category-level base mesh.
\end{itemize}
Note that, in addition to a 3D mesh, FoundationPose requires a ground-truth segmentation mask and a depth image as input, making it substantially more demanding at inference than the RGB-only MegaPose. Results are reported in Table~\ref{tab:benchmark_eval}.

The gap between GT-mesh and base-mesh configurations is striking and
consistent across both methods. When provided with the ground-truth
instance geometry, MegaPose achieves an average ADD-S of
\SI{11.0}{\milli\meter}, but this degrades to
\SI{66.6}{\milli\meter} with the base mesh — a $6\times$
deterioration. Foundation\-Pose exhibits a similar trend, rising from
\SI{20.4}{\milli\meter} (GT) to \SI{29.6}{\milli\meter} (Base) in
average ADD-S, a $1.5\times$ increase. The degradation is most severe
for categories with high geometric variability: for pumpkin,
MegaPose's ADD-S jumps from \SI{11.7}{\milli\meter} to
\SI{93.9}{\milli\meter} ($8\times$), while for carrot it rises from
\SI{24.3}{\milli\meter} to \SI{85.3}{\milli\meter} ($3.5\times$).

These results demonstrate that, for existing methods, access to the
exact instance geometry is a decisive factor in pose accuracy. In
practical agricultural settings, where per-instance 3D scans are
unavailable, this reliance on ground-truth meshes represents a
fundamental limitation. This motivates the need for methods that
explicitly model per-instance shape deformations from the category
prior, rather than assuming a fixed template — the approach taken by
SEED, which we evaluate next.
We also note that MegaPose (GT) outperforms FoundationPose (GT) on
most categories despite lacking depth input, which we believe stems
from subtle RGB-D misalignment in the hardware-aligned depth stream.

\begin{figure}[t]
    \centering
    \includegraphics[width=0.48\textwidth]{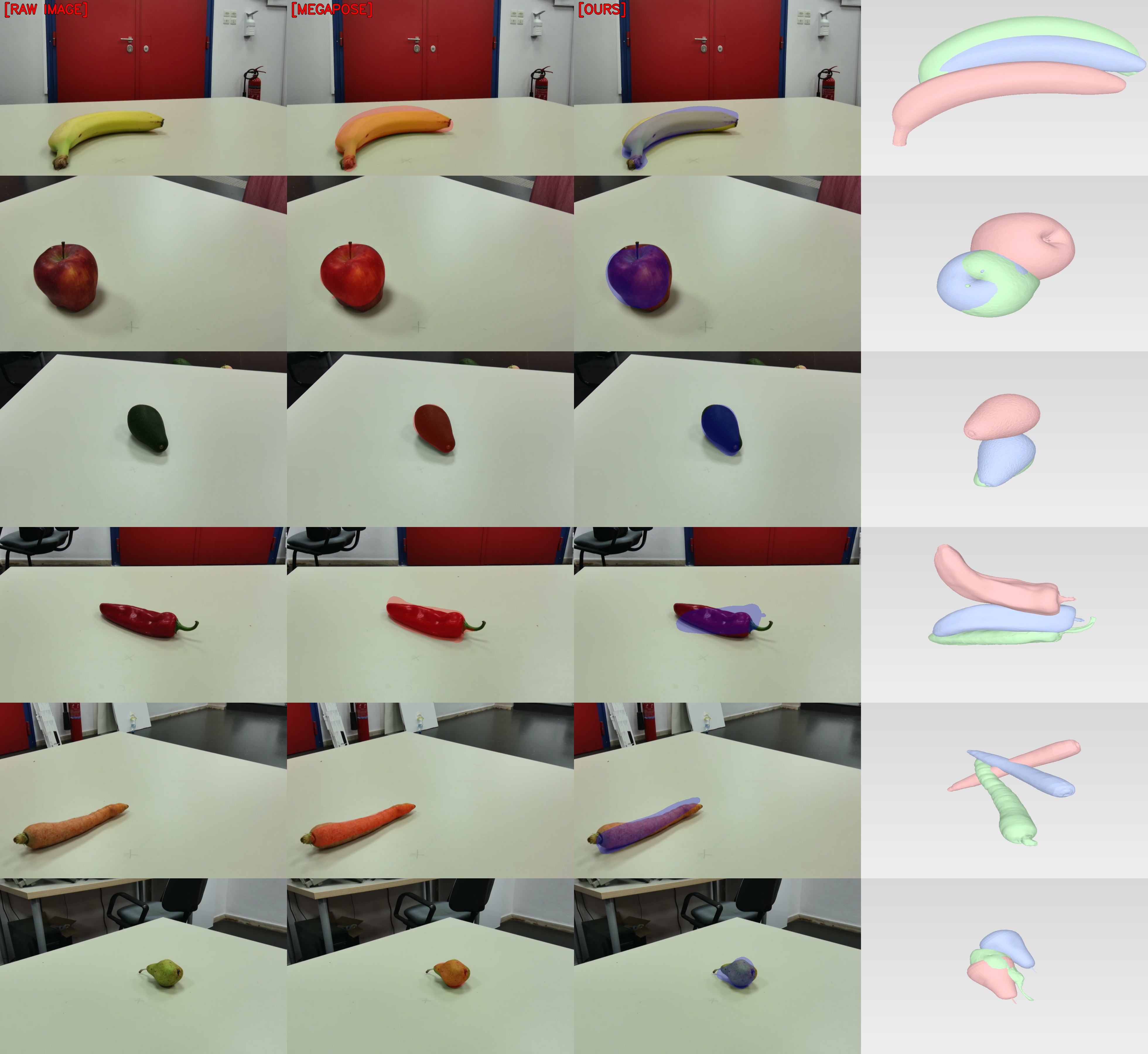}
\caption{Qualitative results on the PEAR benchmark. From left to right: input RGB image, MegaPose~\cite{labbe2022megapose} prediction overlaid with the category base mesh, SEED prediction overlaid with the estimated deformed mesh, and 3D meshes comparison from another viewpoint (green: ground truth, red: MegaPose, blue: ours). SEED produces closer alignment to the ground-truth geometry across categories.}
\label{fig:qualitative}
\vspace{-0.5cm}
\end{figure}

\subsection{SEED Evaluation}
\label{subsec:seed_eval}

Having established the severity of the deformation gap, we now
evaluate whether explicit deformation modeling can recover part of
this lost accuracy in a purely RGB setting. We compare SEED against
MegaPose~\cite{labbe2022megapose} with the category-level base mesh
(MP-Base), as this is the only comparable RGB-only configuration that
does not require depth input or a per-instance 3D scan at inference.
Results are reported in Table~\ref{tab:seed_eval}.

Under identical operating conditions — RGB-only input with no
per-instance geometry — SEED outperforms MegaPose (Base) on 6 out of 8
categories, improving the average ADD-S from
\SI{66.6}{\milli\meter} to \SI{57.1}{\milli\meter}. The gains are
most visible for categories with high shape variability such as
avocado, banana, and carrot, where the base mesh is a particularly
poor approximation of the true instance geometry. On pear and lemon,
MegaPose retains an advantage, as the near-spherical shape of these
categories makes the base mesh already a reasonable approximation.
While SEED does not yet match GT-mesh performance, these results
suggest that explicit deformation modeling is a promising direction for
closing the gap in purely RGB settings.

\subsection{Ablation Study}





Table~\ref{tab:ablation_multi} compares training a separate
single-category model per produce type against our unified
multi-category model. Joint training consistently improves performance
across all three evaluated categories. 

\begin{table}[!htbp]
\centering
\caption{Ablation study comparing single- and multi-category training. Multi-category training achieves the lowest error (mm) across all evaluated agrifoods.}
\label{tab:ablation_multi}
\begin{tabular}{llccc}
\toprule
\textbf{Object} & \textbf{Training} & \textbf{ADD} $\downarrow$ & \textbf{ADD-S} $\downarrow$ & \textbf{Ch.} $\downarrow$ \\
\midrule
\multirow{2}{*}{Banana}
 & Single-model & 74.5  & 48.3 & 49.2 \\
 & Multi-model  & \textbf{67.4} & \textbf{41.6} & \textbf{42.4} \\
\midrule
\multirow{2}{*}{Apple}
 & Single-model & 74.0  & 33.4 & 31.8 \\
 & Multi-model  & \textbf{62.0} & \textbf{23.6} & \textbf{22.9} \\
\midrule
\multirow{2}{*}{Avocado}
 & Single-model & 60.8  & 28.8 & 29.6 \\
 & Multi-model  & \textbf{54.7} & \textbf{25.3} & \textbf{25.1} \\
\midrule
\multirow{2}{*}{Pumpkin}
 & Single-model & 217.2 & 95.8 & 96.1 \\
 & Multi-model  & \textbf{216.2} & \textbf{91.8} & \textbf{102.1} \\
\midrule
\multirow{2}{*}{Pear}
 & Single-model & 149.4 & 112.0 & 116.3 \\
 & Multi-model  & \textbf{124.5} & \textbf{89.7} & \textbf{92.2} \\
\midrule
\multirow{2}{*}{Lemon}
 & Single-model & 120.9 & 90.7 & 90.6 \\
 & Multi-model  & \textbf{100.6} & \textbf{68.7} & \textbf{68.9} \\
\midrule
\multirow{2}{*}{Long Pepper}
 & Single-model & 106.8 & 77.3 & 79.9 \\
 & Multi-model  & \textbf{84.8} & \textbf{56.2} & \textbf{58.6} \\
\midrule
\multirow{2}{*}{Carrot}
 & Single-model & 91.9 & 72.8 & 71.5 \\
 & Multi-model  & \textbf{78.3} & \textbf{60.1} & \textbf{60.6} \\
\bottomrule
\end{tabular}
\vspace{-0.4cm}
\end{table}

\subsection{Qualitative Evaluation}

Fig.~\ref{fig:qualitative} presents qualitative results 
on a representative subset of SEED. For each 
example we show the input RGB image, the predictions of 
each method overlaid on the image, and the estimated 3D 
mesh from a novel viewpoint. In the 3D column, green 
indicates the ground-truth mesh, blue is SEED, and red 
is MegaPose. As can be observed, SEED produces 
more accurate pose and shape estimates across all categories.

\section{Conclusion}

We presented PEAR, the first benchmark providing joint 6D pose and 3D deformation ground truth for agricultural produce, and SEED, a unified RGB-only framework for simultaneous pose and deformation estimation across multiple categories. Our evaluation shows that state-of-the-art methods can suffer up to a 6× drop in accuracy when true instance geometry is unavailable, indicating that shape variability remains a key bottleneck for robotic perception and manipulation. SEED demonstrates that explicit lattice-based deformation modeling can recover a portion of this lost accuracy, outperforming MegaPose on 6 out of 8 categories under identical RGB-only conditions. Future work will explore richer deformation representations, integration with grasp planning, and extension to in-field conditions. 
\bibliographystyle{IEEEtran}
\bibliography{references}

\end{document}